\documentclass{article}
\usepackage{spconf,amsmath,epsfig}
\usepackage{graphicx}
\usepackage{amsmath,amssymb} 
\usepackage{color}
\usepackage{floatrow}
\usepackage{multirow}
\usepackage{lipsum}
\newfloatcommand{capbtabbox}{table}[][\FBwidth]

\newcommand\blfootnote[1]{%
\begingroup
\renewcommand\thefootnote{}\footnote{#1}%
\addtocounter{footnote}{-1}%
\endgroup
}

\begin{document}\sloppy

\def\x{{\mathbf x}}
\def\L{{\cal L}}

\title{Relational Network for Skeleton-Based Action Recognition}
%
\name{Wu Zheng\textsuperscript{1,2,4$\dagger$}, Lin Li\textsuperscript{1,2,4$\dagger$}, Zhaoxiang Zhang\textsuperscript{1,2,3,4*}, Yan Huang\textsuperscript{1,2,4}, Liang Wang\textsuperscript{1,2,4}}
\address{
$^1$Center for Research on Intelligent Perception and Computing, CASIA, China \\
$^2$National Laboratory of Pattern Recognition, CASIA, China \\
$^3$CAS Center for Excellence in Brain Science and Intelligence Technology, China\\
$^4$University of Chinese Academy of Sciences,  China\\
\{zheng-w10@foxmail.com, zhaoxiang.zhang@ia.ac.cn\}}

\maketitle

\begin{abstract}
\blfootnote{$\dagger$ Equal contribution}
\blfootnote{$*$ Corresponding author}
With the fast development of effective and low-cost human skeleton capture systems, skeleton-based action recognition has attracted much attention recently. Most existing methods use Convolutional Neural Network (CNN) and Recurrent Neural Network (RNN) to extract spatio-temporal information embedded in the skeleton sequences for action recognition. However, these approaches are limited in the ability of relational modeling in a single skeleton, due to the loss of important structural information when converting the raw skeleton data to adapt to the input format of CNN or RNN. In this paper, we propose an Attentional Recurrent Relational Network-LSTM (ARRN-LSTM) to simultaneously model spatial configurations and temporal dynamics in skeletons for action recognition. We introduce the Recurrent Relational Network to learn the spatial features in a single skeleton, followed by a multi-layer LSTM to learn the temporal features in the skeleton sequences. Between the two modules, we design an adaptive attentional module to focus attention on the most discriminative parts in the single skeleton. To exploit the complementarity from different geometries in the skeleton for sufficient relational modeling, we design a two-stream architecture to learn the structural features among joints and lines simultaneously. Extensive experiments are conducted on several popular skeleton datasets and the results show that the proposed approach achieves better results than most mainstream methods.
\end{abstract}
\section{Introduction}
\label{sec:intro}

Action recognition provides a reasonable approach for video understanding and is under great demand, especially in the domains of intelligent surveillance and human-computer interaction. Traditional approaches are mainly based on the modeling of appearance and optical flow. However, the noise interference in RGB video dramatically obstructs the extraction of high-level features for action recognition.

Benefited from the advent of affordable depth sensors and efficient algorithms, dynamic human skeleton becomes an available and effective modality for action recognition. Meanwhile, compared with RGB video, the characteristics of high-level representation and robustness to viewpoints,
\begin{figure}
\centering
\includegraphics[width=8cm]{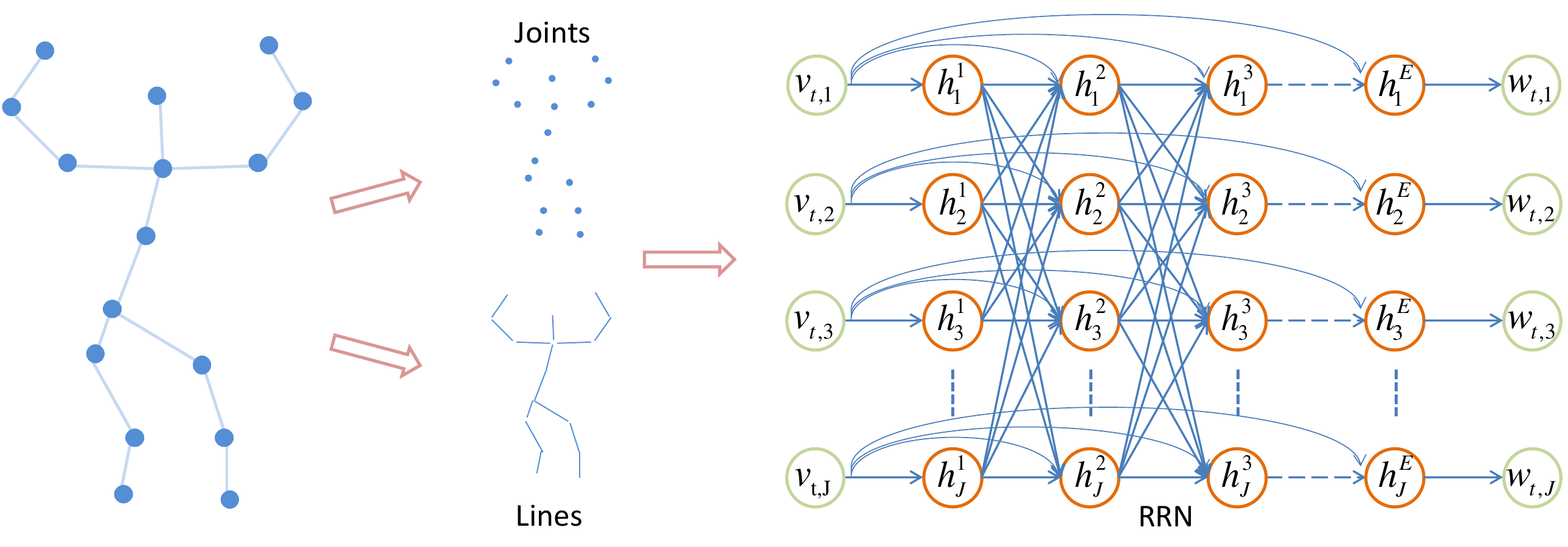}
\caption{The expanded structure of Recurrent Relational Network (RRN), which is used to learn the spatial pattern in the single skeleton frame by modeling joints and lines separately.}
\label{fig1}
\end{figure}
appearances and background noise make skeletons have advantages in action recognition. As a result, many early skeleton-based methods were proposed and have showed encouraging improvements, such as \cite{wang2012mining,vemulapalli2014human,zanfir2013moving}. However, these approaches were significantly limited in either the lack of exploring spatial structures \cite{wang2012mining} or the dependence for hand-crafted features to analyze the spatial patterns \cite{vemulapalli2014human,zanfir2013moving}.

Recently, various deep learning based methods have been proposed to conduct skeleton-based action recognition. In general, these approaches are mainly based on CNN and RNN for capturing spatio-temporal information in skeletons. Specifically, the CNN-based methods utilize the powerful representation ability of CNN and achieve better performances than those hand-crafted feature based methods. And the RNN-based models have shown great advantages in capturing temporal dynamics in sequential skeletons. However, CNNs usually lose important structural information in the process of encoding skeletons into spatial-temporal images \cite{hou2016skeleton,du2015skeleton,liu2017two}, and RNNs have the same weakness when learn the spatial features in a single skeleton \cite{du2015hierarchical,song2017end,shahroudy2016ntu,liu2016spatio}. Thus, when converting the raw skeleton data to match with the CNN or RNN input format, the destruction of the original structures among the skeleton joints and lines leads to the difficulty in extracting robust spatial features in a single skeleton, which remains the main weakness of these frameworks.

In this paper, we propose an Attentional Recurrent Relational Network \cite{palm2017recurrent}-LSTM (ARRN-LSTM) to model temporal dynamics and spatial configurations in skeletons for action recognition. Our approach is based on a two-stream architecture to learn sufficient relational information from both joints and lines in the skeleton. In each stream, we use the Recurrent Relational Network to learn the spatial patterns in a single skeleton and exploit a multi-layer LSTM to extract temporal information in skeleton sequences. Between the two modules, we design an adaptive attentional module for focusing on potential discriminative parts of a skeleton towards a certain action. Compared with other graph networks \cite{li2018spatio,yan2018spatial}, we believe Recurrent Relational Network is a better framework for learning spatial information in a single skeleton, as depicted in Fig.\ref{fig1}, because it can ensure the flexible flow of information and build long-range dependencies among all joints or lines, which is important for learning robust features from graph structure data. Overall, our contributions can be summarized as follows:
\begin{itemize}
\item We introduce the Recurrent Relational Network to the domain of skeleton-based action recognition and prove that it is a very good framework to learn the spatial feature in the single skeleton.
\item We design an organic framework, the two-stream ARRN-LSTM, to conduct skeleton-based action recognition, and achieve better results than most mainstream methods on popular skeleton datasets.
\end{itemize}

\section{Related Work}

\subsection{Relational Network}
Santoro et al. \cite{santoro2017simple} propose a simple plug-and-play neural network module for relational reasoning. With this module, a neural network gains the ability of handling unstructured inputs and inferring their hidden relationship, which achieves state-of-the-art results on visual question answering datasets. Based on this work, Palm et al. \cite{palm2017recurrent} propose the Recurrent Relational Networks for complex relational reasoning, such as learning an iterative strategy to solve Sudoku.

\subsection{Skeleton-based Action Recognition with Deep Networks}
To utilize the powerful representation ability of CNN, skeletons are usually encoded into spatial-temporal images to fit the inputs of CNN. Hou et al. \cite{hou2016skeleton} accumulate the raw skeleton frames directly and encode the color based on temporal information. Liu et al. \cite{liu2017two} exploit the 3DCNN to extract spatio-temporal features for avoiding the loss of information in projecting process. Du et al. \cite{du2015skeleton} divide the joints into five main parts according to human physical structures (four limbs and one trunk) and take the 3D coordinates of joints as 3 channels of RGB image. Specifically, Yan et al. \cite{yan2018spatial} use the Graph Convolutinal Network to form hierarchical representation of skeletons and achieve good results.

RNN is good at processing sequential data due to the extraordinary ability of capturing structural information in sequences. Du et al. \cite{du2015hierarchical} divide the human skeleton into five parts according to human physical structure and separately feed them into different RNNs. Song et al. \cite{song2017end} modify RNN to design an attentional module and use multi-layer LSTM to learn spatial and temporal information. Shahroudy et al. \cite{shahroudy2016ntu} propose a Part-Aware LSTM unit that builds full connections between all the memory cells and all the input features for acquiring richer information. Liu et al. \cite{liu2016spatio} transform the joints in the form of tree structure based on traversal and propose a spatio-temporal LSTM framework to learn spatio-temporal information in joint sequences.

\section{Proposed Method}
\subsection{Pipeline Overview}

\begin{figure*}
\centering
\includegraphics[width=18cm]{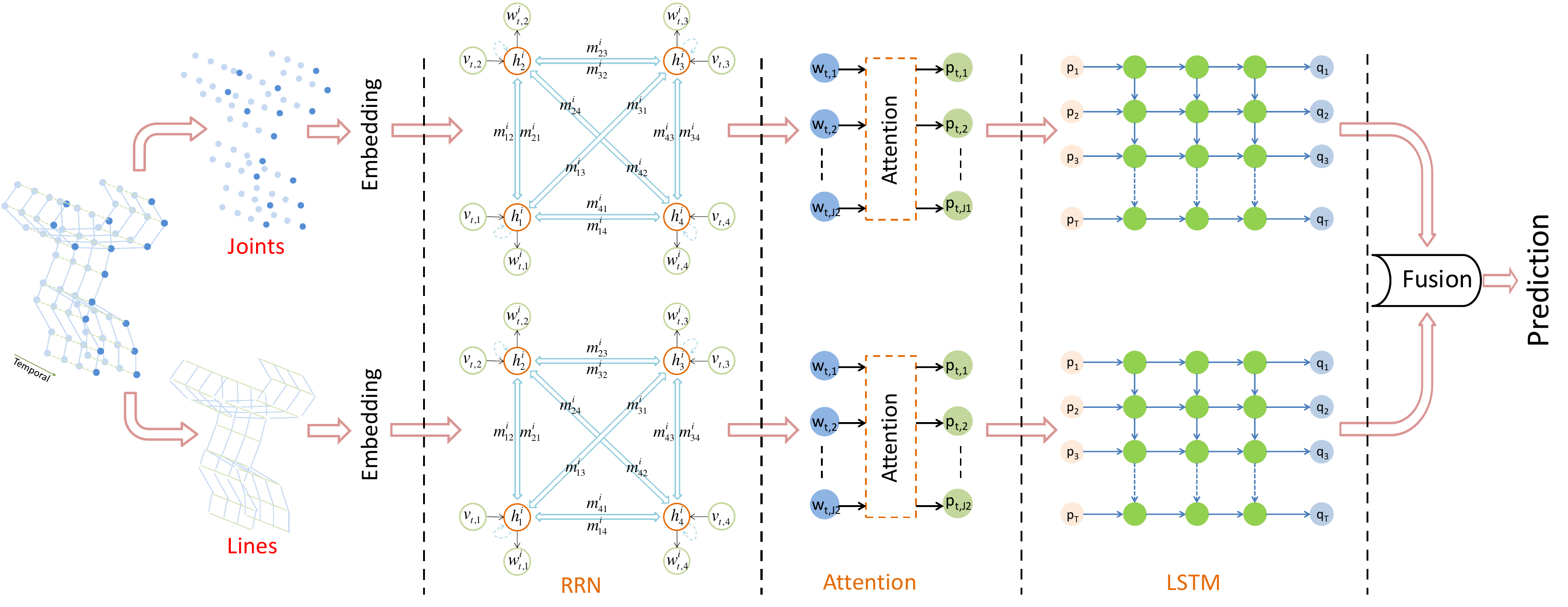}
\caption{Framework of the proposed two-stream ARRN-LSTM model. It is recommended to view the digital version.}
\label{fig2}
\end{figure*}

The pipeline of our framework is depicted in Fig.\ref{fig2}. Our framework consists of two streams for learning structural features from joints and lines separately. In each stream, an embedding operation is first performed on each joint or line to increase its dimension. Then, the embedding results of joints or lines are sent to RRN for capturing the spatial patterns in the single skeleton. To focus more attention on potential discriminative parts in the skeleton, we generate a learnable mask and then use it to perform point-wise multiplication with the outputs of RRN. After that, we use a multi-layer LSTM to learn temporal features in skeleton sequences. Finally, we take the weighted average operation as our fusion strategy to combine the predictions from both streams.

\subsection{The Construction of the Two-Stream ARRN-LSTM}
\subsubsection{Embedding} The raw skeleton is defined by a fixed number of joints in the form of 3D coordinates, with the number denoted as $J$. To improve the representation of joints and strengthen the discrimination among them, we use a fully-connected layer to map the 3D coordinates to a high dimensional space. Thus, given a joint $c_{t,i} = (c_{t,i}^{x},c_{t,i}^{y},c_{t,i}^{z})$ that means the 3D coordinate of the $i$-th joint in the $t$-th frame, the $M$-dimensional embedding result $v_{t,i}$ is:

\begin{equation}
    \begin{matrix}
        $$v_{t,i} = {\rm Emb}(c_{t,i}) = {\rm Emb}(c_{t,i}^{x},c_{t,i}^{y},c_{t,i}^{z})$$\\
    \end{matrix}
    \label{eq1}
\end{equation}

With the exception of original joints, we believe that the lines between pair-wise joints are also important geometric structures in the skeleton and contain rich structural or relational information. Specifically, joints emphasize the absolute position, which can figure out discriminative moving parts of body in the action by analyzing the distribution or local density of joints. While lines emphasize relative position, which can build angles to help figure out specific poses and actions. Thus, there exists potential complementarity for action recognition between both geometric structures. We calculate lines between $c_{t,i}$ and other joints as:
\begin{equation}
    \begin{matrix}
        $$l_{t,i} = (c_{t,i}^{x} - c_{t,1}^{x},c_{t,i}^{y} - c_{t,1}^{y},c_{t,i}^{z} - c_{t,1}^{z},c_{t,i}^{x} - c_{t,2}^{x}...,
        \\c_{t,i}^{x} - c_{t,J}^{x},c_{t,i}^{y} - c_{t,J}^{y},c_{t,i}^{z} - c_{t,J}^{z})
        $$
    \end{matrix}
    \label{eq1_f=ma}
\end{equation}
The $l_{t,i}$ denotes all lines that connect the joint $c_{t,i}$ and all the other joints, but these distances exclude the one between $c_{t,i}$ and itself. Thus, if the dimension of one joint is $3$, the dimension of the corresponding lines is $3\times(J-1)$. Similarly, the line embedding process is:
\begin{equation}
    \begin{matrix}
        $$v_{t,i} = {\rm Emb}(l_{t,i}) = {\rm Emb}(l_{t,i}^{1},l_{t,i}^{2},l_{t,i}^{3}...,l_{t,i}^{3\times(J-1)})$$\\
    \end{matrix}
    \label{eq1}
\end{equation}
Because the output dimensions of both joint and line embeddings are equal, we use the same symbol $v_{t,i}$ to denote the embedding result of lines $l_{t,i}$ in following expressions.

\subsubsection{Recurrent Relational Network}

In the domain of skeleton-based action recognition, some graph networks based frameworks \cite{li2018spatio,yan2018spatial} have been proposed recently and achieved substantial improvements than mainstream methods. While these frameworks applied the notion of receptive fields to the graph structure and extracted features only from a small adjacent range. Compared with them, we believe Recurrent Relational Network is a better framework for learning spatial features in a single skeleton, because it can ensure the flexible flow of information among joints and build long-range dependencies among them, which could help to learn more robust features from graph structure data. Besides, the implementation of RRN is much easier than previous graph networks.

Specifically, we use all joints or lines from a single skeleton to feed the RRN. In the process, each joint or line embedding $v_{t,i}$ will be the input to one node of RRN and the number of nodes in RRN is $J$, which can be denoted as:
\begin{equation}
    \begin{matrix}
        $$w_{t} = {\rm Rrn}(v_{t\cdot})= {\rm Rrn}(v_{t,1},v_{t,2},v_{t,3}...,v_{t,J})\\
    \end{matrix}
    \label{eq1}
\end{equation}
As for the detailed process, if we denote the states of node $i$ and $j$ in the $e$-th iteration of RRN as $h_{i}^{e}$ and $h_{j}^{e}$, we can define the information flowing from node $j$ to node $i$ as:
\begin{equation}
    \begin{matrix}
        $$m_{i,j}^{e} = f(h_{i}^{e-1}$,$h_{j}^{e-1})\\
    \end{matrix}
    \label{eq1}
\end{equation}
$f$ is a message function. The messages from all neighbouring nodes to joint $i$ are then summed up as:
\begin{equation}
    \begin{matrix}
        $$m_{i\cdot}^{e} =  \sum_{j\in N(i)}m_{i,j}^{e}\\
    \end{matrix}
    \label{eq1}
\end{equation}
Then the ouput of node $i$ can be updated by a trainable node function $g$ as:
\begin{equation}
    \begin{matrix}
        $$h_{i}^{e} =  g(h_{i}^{e-1},v_{t,i},m_{i\cdot}^{e})\\
    \end{matrix}
    \label{eq1}
\end{equation}
Given the number of iterations in RRN as $E$, the output of node $i$ after the final iteration can be expressed as:
\begin{equation}
    \begin{matrix}
        $$w_{t,i} =  h_{i}^{E}\\
    \end{matrix}
    \label{eq1}
\end{equation}
At this point, we calculate the relational modeling results of the joints or lines in a skeleton as $w_{t} = (w_{t,1},w_{t,2},w_{t,3}...,w_{t,J})$, in the dimension of $(J,M)$.

\subsubsection{Attentional Module}
For a certain action, humans usually recognize it by focusing the most discriminative parts. For example, the action of kicking can be identified through the legs, while the drinking action can be recognized by the arms. However, some different actions, such as flipping and reading a book, cannot be distinguished until the subtle differences in the hand part are identified. Thus, we design an attentional module to address these problems, with the module following the RRN. Specifically, we first generate a learnable mask, it could be a random vector of size $J$, and then we use it to perform point-wise multiplication with the node outputs of RRN. After that, we use a fully-connected layer to reduce the product dimension to avoid overfitting in following multi-layer LSTM. We express the process as follows:
\begin{equation}
    \begin{matrix}
        $$ p_{t} = {\rm Att}(w_{t}) = {\rm Fc}(m_{1}\cdot w_{t,1}, m_{2}\cdot w_{t,2}..., m_{J}\cdot w_{t,J})\\
    \end{matrix}
    \label{eq1}
\end{equation}
The mask $m=(m_{1},m_{2},...,m_{J})$ is a $J$-dimensional vector and can be trained with back-propagation algorithm, which aims to emphasize the impacts of some joints or lines while neglect other unimportant ones by allocating different weights. The output $p_{t}$ can expressed as $ (p_{t,1},p_{t,2},p_{t,3}...,p_{t,J})$.

\subsubsection{Multi-Layer LSTM}
Following RRN, we exploit a multi-layer LSTM to learn the temporal dynamics in skeleton sequences. We first concatenate all joints or lines features from a single skeleton as the input of one cell in LSTM, and the number of cells in LSTM is equal to the skeleton sequence length $T$, so all frames in a skeleton sequence can exchange information with internal connections in LSTM. This process is denoted as follows:
\begin{equation}
    \begin{matrix}
        $$(q_{1},q_{2},q_{3}...,q_{T}) = {\rm Multi\_lstm}(p_{1},p_{2},p_{3}...,p_{T})\\
    \end{matrix}
    \label{eq1}
\end{equation}
The number of layers in multi-layer LSTM  is $H$ and the dimension of $q_{t}$ is $h$.

\subsubsection{Score Fusion}
To get the final prediction from each stream, we connect a fully-connected layer after multi-layer LSTM to map the extracted spatial-temporal features to the categories of size $K$. Then we run a softmax operation on the output to obtain the predicted probabilities. We take joint stream as an example:
\begin{equation}
    \begin{matrix}
        $$y_{j} = {\rm softmax}({\rm Fc}(q_{1},q_{2},q_{3}...,q_{T}))\\
    \end{matrix}
    \label{eq1}
\end{equation}

To exploit the complementarity between joints and lines, we take the weighted average as the fusion strategy and get the final prediction. We use $y_{j}$ and $y_{l}$ to denote the scores of joint and line stream, respectively. The final prediction $y$ is:
\begin{equation}
    \begin{matrix}
        $$y = \alpha \cdot y_{j} + \beta \cdot y_{l}\\
    \end{matrix}
    \label{eq1}
\end{equation}
$\alpha$ and $\beta$ are relative weights of the two stream predictions and their sum is equal to 1.

\section{Experiments and Results}
In our experiments, firstly we perform detailed ablation study to validate our framework, showing the results in Tables \ref{table5} and \ref{table4}. Then we train the proposed two-stream Attentional RRN-LSTM model and test it on NTU RGB+D, Florence 3D, and MSRAction3D datasets, with results and comparisons shown in Tables \ref{table1}, \ref{table2} and  \ref{table3} separately.

\subsection{Implementation Details}
We normalize the joint coordinators by subtracting the average value of the 5 joints close to the hip joint. The lines are calculated according to the normalized joints. We perform zero padding on videos with frames less than $T$ and random sampling on videos with frames more than $T$ to fix all videos as $T$ frames. According to the differences of frame numbers in different datasets, we set $T=100,25,20$ for NTU-RGBD, Florrence3D and MSRAction3D, respectively. Besides, we conduct the embedding with a fully-connected layer and set the output size $M=50,20,20$ for the three datasets based on cross-validation. The following RRN executes $E=5$ iterations per frame with each node function $g$ realized by an GRU unit and the message function $f$ constructed with 3 fully-connected layers. The attentional module is built with a trainable mask $m$ and a fully-connected layer for reducing the product to a 256-dimensional vector. We use $H=3$ layers LSTM to extract temporal information in skeleton sequences, with input and output set as a 256-dimensional and $h=512$-dimensional vector separately. The mid-layer LSTM has an output size of 512,256,256 for NTU-RGBD, Florrence3D and MSRAction3D, respectively. Both $\alpha$ and $\beta$ are set as the optimal values based on validation set in our experiments. We use Stochastic Gradient Descent to train our model from scratch on NTU-RGBD and set the initial learning rate as 0.01, we multiply the learning rate with 0.1 when the accuracy gets saturated. On other datasets, we use Adam optimizer to train our model from scratch. Our model is trained on a NVIDIA TITAN X GPU with PyTorch.

\subsection{Datasets}
\textbf{NTU RGB+D Dataset.} This is the most popular and largest depth-based skeleton action recognition dataset currently, with more than 56 thousand video samples and 4 million frames collected from 40 different subjects. It consists of 60 different action classes. We evaluate our model according to the metrics proposed in \cite{shahroudy2016ntu}, including Cross-Subject (CS) and Cross-View (CV) evaluations.

\textbf{Florence 3D.} This dataset includes 215 action sequences performed by 10 subjects for 2 to 3 times. It is made up of 9 activities and each skeleton is represented by 15 joints. The difficulty of this dataset lies in its similarities between actions, such as drinking from a bottle, answering phone and reading watch. We follow the standard metric, i.e., leave-one-subject-out cross validation, to evaluate our model.

\textbf{MSRAction3D.} This dataset contains 20 actions performed by seven subjects for three times, which totally consists of 4020 action samples. The dataset is divided into three subsets and each subset has 8 actions. In each subset, the samples of subjects 1, 3, 5, 7, 9 are used for training while the samples of subjects 2, 4, 6, 8, 10 are used for testing. Final accuracy is calculated as average accuracies of three subsets.

\subsection{Ablation Study}
We examine the effectiveness of the proposed ARRN-LSTM framework and study the impact of each part by ablation study in this section, with results on NTU shown in Table \ref{table5} and \ref{table4}.

\begin{table}\small
\setlength{\tabcolsep}{4mm}{ 
\begin{center}
\caption{Comparisons with Baselines on NTU RGB+D.}
\begin{tabular}{cccccc}
\hline
{Methods}  &CS      &CV \\
\hline
2-Layer LSTM\cite{shahroudy2016ntu}  &60.7  &67.3  \\
2-Layer P-LSTM\cite{shahroudy2016ntu} &62.9 &70.3  \\
MT-3D-CNN\cite{liu2017two}  &66.9    &72.6   \\
STA-LSTM\cite{song2017end}   &73.4    &81.2   \\
VA-LSTM\cite{zhang2017view}    &79.4    &87.6   \\
\hline
Joint-ARRN-LSTM & \textbf{79.6}   &\textbf{87.8}   \\
\hline
\label{table5}
\end{tabular}
\end{center}
}
\vspace{-1.2cm}
\end{table}

\begin{table}\small
\setlength{\tabcolsep}{4mm}{ 
\begin{center}
\caption{Results of Ablation Study on NTU RGB+D.}
\begin{tabular}{ccc}
\hline
Model        &CS             &CV \\
\hline
Line-RRN-LSTM &74.5            &83.3 \\
Joint-RRN-LSTM &74.6            &83.1\\
Line-ARRN-LSTM  &76.4               &87.2\\
Joint-ARRN-LSTM &79.6               &87.8\\
Two-Stream RRN-LSTM &77.6            &84.2\\
Two-Stream ARRN-LSTM &\textbf{81.8}               &\textbf{89.6}\\
\hline
\label{table4}
\end{tabular}
\end{center}
}
\vspace{-1.2cm}
\end{table}

To illustrate the effectiveness of RRN in learning spatial features in single skeleton, we compare the results of our joint stream with several baselines in Table \ref{table5}, and these methods also learn features from joints. Our method uses RRN and multi-layer LSTM to learn spatial and temporal information separately. Differently, 2-layer LSTM and 2-layer P-LSTM use pure LSTM or its variant to extract spatio-temporal features from joints, and our method outperforms it substantially. Furthermore, MT-3D-CNN \cite{liu2017two} uses CNN to build a two-stream model for learning spatial and temporal features from skeletons, while both STA-LSTM \cite{song2017end} and VA-LSTM \cite{zhang2017view} use multi-layer LSTM as backbone to learn spatial and temporal information from joints. But our method achieves much better results than these baselines, which validates that RRN can model spatial features in single skeleton effectively.

From Table \ref{table4}, we analyze the performance of each stream and attentional module. Each stream ARRN-LSTM can achieve relatively good performance, which validates the effectiveness of the basic ARRN-LSTM. For attentional module, it increases the accuracy by $2\sim5$ points under both metrics, validating the insight of the attention mechanism. Furthermore, the combination of both streams can result in an increase of $2\sim3$ points in final accuracy, proving the complementarity between joints and lines. Finally, the overall framework increases the accuracy by almost $6\sim7$ points, which proves the reasonability and effectiveness of our framework.

For embedding layer, we cannot ensure the convergence of the framework if we remove it, so we do not show the ablation study of this part. For fusion of two streams, we tried early fusion and several operations like max and multiply, but they are hard to converge or achieve good performances.

\begin{table}{
\begin{center}
\caption{Comparison of Accuracies on NTU RGB+D.}
\begin{tabular}{ccc}
\hline
Methods                                  & CS(\%) & CV(\%)     \\
\hline
Lie Group   \cite{vemulapalli2014human}    &50.1  &52.8  \\
Deep LSTM \cite{shahroudy2016ntu}             &60.7  &67.3   \\
PA-LSTM \cite{shahroudy2016ntu}               &62.9  &70.3   \\
ST-LSTM+TS  \cite{liu2016spatio}              &69.2  &77.7   \\
\textbf{ST-NBMIM} \cite{weng2018discriminative} &80.0 &84.2 \\
\textbf{Deep STGCK} \cite{li2018spatio}         &74.9 &86.3   \\
VA-LSTM \cite{zhang2017view}           &79.4  &87.6   \\
C-CNN + MTLN \cite{ke2017new}     &79.6 &84.8   \\
\textbf{ST-GCN} \cite{yan2018spatial}         &81.5  &88.3    \\
\hline
ARRN-LSTM           &\textbf{81.8} &\textbf{89.6}   \\
\hline
\label{table1}
\end{tabular}
\end{center}
}
\vspace{-1.2cm}
\end{table}

\begin{table}
\begin{center}
\caption{Comparison of Accuracies on Florence 3D.}
\begin{tabular}{cc}
  \hline
    Methods            & Accuracy(\%)\\
    \hline
    Lie Group \cite{vemulapalli2014human} &90.88\\
    Graph-Based \cite{wang2016graph} &91.63\\
    P-LSTM \cite{shahroudy2016ntu} &95.35\\
    STGCK \cite{li2018spatio} &97.67\\
    \textbf{Deep STGCK} \cite{li2018spatio} &\textbf{99.07}\\
    \hline
    ARRN-LSTM           &98.52\\
    \hline
    \label{table2}
\end{tabular}
\end{center}
\vspace{-1.2cm}
\end{table}

\subsection{Comparisons with Mainstream Methods}

\textbf{NTU RGB+D Dataset.} We compare our ARRN-LSTM model with mainstream methods on this dataset in Table \ref{table1}, it is clear that our ARRN-LSTM method achieves better results than those CNN or RNN based methods. Specifically, ST-GCN \cite{yan2018spatial} takes the graph convolution to learn the spatial and temporal features in skeleton sequences, while our method takes the RRN to perform relational modeling in the single skeleton and uses a multi-layer LSTM to obtain the temporal information in skeleton sequences. The results could prove the strong ability of the RRN in modeling spatial features in single skeleton and the effectiveness of the whole framework.

\begin{table}
\begin{center}
\caption{Comparison of Accuracies on MSRAction3D.}
\begin{tabular}{cc}
 \hline
    Methods                                   & Accuracy(\%)\\
    \hline
    Lie Group \cite{vemulapalli2014human}      &92.5\\
    HBRNN \cite{du2015hierarchical}            &94.5\\
    ST-LSTM \cite{liu2016spatio}               &94.8\\
    Graph-Based \cite{wang2016graph}           &94.8\\
    ST-NBNN \cite{weng2018discriminative}      &94.8\\
    \textbf{ST-NBMIM} \cite{weng2018discriminative}     &\textbf{95.3}\\
    \hline
    ARRN-LSTM                                  &95.0\\
    \hline
    \label{table3}
\end{tabular}
\end{center}
\vspace{-1.2cm}
\end{table}

\textbf{Florence 3D.} As shown in Table \ref{table2}, the proposed ARRN-LSTM framework is superior to most mainstream methods that are based on LSTM, CNN and traditional algorithms. Compared with the state-of-the-art Deep STGCK \cite{li2018spatio}, the performance of our model is very close to its result. Our model is also based on a graph network like Deep STGCK, but the multi-layer LSTM make model complexity much larger than Deep STGCK, making it easy to be overfitting on small datasets.

\textbf{MSRAction3D.} As shown in Table \ref{table3}, our method achieves $95.0\%$ accuracy and outperforms most mainstream methods, and the result is also competitive with the state-of-the-art method ST-NBMIM \cite{weng2018discriminative}. Although our framework suffers overfitting on small datasets again, the result could still validate the effectiveness of our method on small datasets.

\section{Conclusion}
In this paper, we introduced the Recurrent Relational Network to the domain of skeleton-based action recognition, and we also designed an organic two-stream ARRN-LSTM framework and achieved better results than most mainstream methods. The experiments results proved the strong modeling ability of RRN in the single skeleton and the effectiveness of the whole framework. However, we believe there still exists possibility of modeling both single skeleton and skeleton sequences with relation network for action recognition, which may provide a brand new perspective to address this problem and a potential direction to achieve further progress.

\section*{Acknowledgement}
This work was supported in part by the National Key {R$\&$D} Program of China (No. 2018YFB1004600), the Beijing Municipal Natural Science Foundation (No. Z181100008918010), the National Natural Science Foundation of China (No. 61836014, No. 61761146004, No. 61773375), and in part by the Microsoft Collaborative Research Project.

{\small
\bibliographystyle{IEEEbib}
\bibliography{icme2019template}
}
\end{document}